\documentclass[a4paper,conference]{IEEEtran}
\usepackage{cite}

\usepackage[T1]{fontenc}

\ifCLASSINFOpdf
   \usepackage[pdftex]{graphicx}
\else
\fi
\usepackage{hhline}
\usepackage{multirow}
\usepackage{colortbl}
\usepackage{xcolor,colortbl}

\definecolor{Gray}{gray}{0.65}
\definecolor{LightCyan}{rgb}{0.88,1,1}

\newcolumntype{a}{>{\columncolor{Gray}}c}

\usepackage{siunitx}
\usepackage{amssymb}

\usepackage{algorithm}
\usepackage{algorithmic}

\usepackage{array}

\ifCLASSOPTIONcompsoc
  \usepackage[caption=false,font=normalsize,labelfont=sf,textfont=sf]{subfig}
\else
  \usepackage[caption=false,font=footnotesize]{subfig}
\fi
\begin{document}
%
\title{Walk the Lines: Object Contour Tracing CNN\\ 
for Contour Completion of Ships}

\author{\IEEEauthorblockN{Andr\'{e} Peter Kelm}
\IEEEauthorblockA{Helmut Schmidt University\\
Department of Signal Processing and Communications\\
22043 Hamburg, Germany\\
Email: andre.kelm@hsu-hh.de}
\and
\IEEEauthorblockN{Udo Z\"olzer}
\IEEEauthorblockA{Helmut Schmidt University\\
Department of Signal Processing and Communications\\
22043 Hamburg, Germany\\
Email: udo.zoelzer@hsu-hh.de}
}


%


\maketitle

\begin{abstract}
We develop a new contour tracing algorithm to enhance the results of the latest object contour detectors. The goal is to achieve a perfectly closed, 1 pixel wide and detailed object contour,
since this type of contour could be analyzed using methods such as Fourier descriptors. Convolutional Neural Networks (CNNs) are rarely used for contour tracing. However, we find CNNs are tailor-made for this task and that’s why we present the Walk the Lines (WtL) algorithm, a standard regression CNN trained to follow object contours. To make the first step, we train the CNN only on ship contours, but the principle is also applicable to other objects. Input data are the image and the associated object contour prediction of the recently published RefineContourNet. The WtL gets a center pixel, which defines an input section and an angle for rotating this section. Ideally, the center pixel moves on the contour, while the angle describes upcoming directional contour changes. The WtL predicts its steps pixelwise in a selfrouting way. To obtain a complete object contour the WtL runs in parallel at different image locations and the traces of its individual paths are summed. In contrast to the comparable Non-Maximum Suppression method, our approach produces connected contours with finer details. Finally, the object contour is binarized under the condition of being closed. In case all procedures work as desired, excellent ship segmentations with high IoUs are produced, showing details such as antennas and ship superstructures that are easily omitted by other segmentation methods.
\end{abstract}


%
\IEEEpeerreviewmaketitle

\section{Introduction}
Complete object contours extracted from an image contain relevant information about the shape of the photographed objects. They are used in many areas of computer vision, like object detection \cite{7783667, 10.1007/978-3-642-10546-3_4}, object recognition \cite{7877440, 5565098} and object tracking \cite{1335457,6682875}. That's why there has always been a great demand for methods that can extract object contours \cite{6376435, 4378938, 6738632}. Recently driven by the success of deep learning methods, the object contour detection has made great progress \cite{CEDN, WeaklySuperObjBound, COB, 10.1007/978-3-030-29888-3_20}. This progress can also be found in the closely related task of semantic edge detection \cite{casenet, seal, steal}, in which the contours are also assigned to object classes such as person, car or dog. The outputs of all these detectors are so-called soft contour maps, and they provide a probability for each pixel to be an object contour.
The mainly used post-processing method, the Non-Maximum Suppression (NMS) \cite{4767851}, breaks the soft contour when thinning out, especially if the contour changes direction. Recently, the NMS has gotten integrated into the end-to-end training of deep networks for improvement \cite{DFF}, but we still see the properties of the NMS resulting in an unconnected object contour.
A quick way to synthesise and analyse them with methods like Fourier descriptors (FDs) is missing, because these normally require complete contours \cite{5008949, Latecki2000ShapeDF}, with few exceptions and loss of performance \cite{5652178}. An easy utilization would be very useful because FDs or similar methods \cite{537644} are extensively researched and used for active contours \cite{6115601}, shape description \cite{7177425}, shape matching \cite{1359759} or identification \cite{8084832}. There is work on contour completion \cite{Guan2015ANC,6247755}, but it should be even more effective not to let the problem arise and to use another post-processing method instead of the NMS. 
For this, we need to improve the soft contour map in such a way that a perfectly closed, 1 pixel wide and detailed binary object contour can be obtained by standard image processing tools in a final step. For the improvement, we develop a new contour tracing algorithm based on Convolutional Neural Network (CNN), because we assume that CNNs are tailor-made for this task. There are some line following approaches based on CNN or Artificial Neural Network (ANN) to support specific robotic tasks \cite{8039205, 8466525}, but overall, CNNs are used rarely for contour tracing. Most methods mainly do not make use of CNNs \cite{ZAMPERONI1982161} and are often intended for binary images \cite{5533002, s16030353, SUZUKI198532}. That’s why we present the Walk the Lines (WtL) algorithm, a standard regression CNN trained to follow object contours in RGB images. To take the first step, we train the WtL only on ship contours, but the principle is applicable to other objects. 

The main advantage of our approach against the NMS is that our WtL contour is closed, connected and a bit more detailed, so that further processing can lead to a complete and detailed binary object contour.


Section 2 describes the new contour tracing algorithm and its training, Section 3 contains its application to generate a complete object contour,
Section 4 will briefly summarize the image processing steps for the object contour binarization,
Section 5 evaluates our results and Section 6 gives a conclusion.
\section{Walk the Lines algorithm}
We propose our method to bridge the gap between promising object contour detector results and a final, highly detailed and closed binary contour. To elaborate details and simultaneously create a connected contour, it is designed as a contour tracing method. 
\subsection{Functionality}
\begin{figure}
\centering
\includegraphics[width=3.35in]{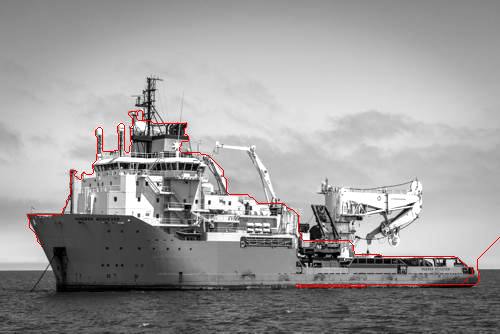}
\caption{A tracer walks the lines from bow to stern and beyond. The line is shown in red and the RGB-image \cite{unsplash} is converted to gray and adjusted in contrast for visualization purposes only.}
\label{onetracer}
\end{figure}

\begin{algorithm}
 \caption{\textit{Walk the Lines}}
 \begin{algorithmic}[1]
 \renewcommand{\algorithmicrequire}{\textbf{Input:}}
 \renewcommand{\algorithmicensure}{\textbf{Output:}}
 \REQUIRE $tensor_{h \times w \times 4}$, $cp$ , $\alpha$
 \ENSURE  $cp_{new}, \alpha_{new}$
 \\ \textit{Initialisation}: $tracer(cp,\alpha)$
  \FOR {$1$ to $x$} 
  \STATE $tensor_{13 \times 13 \times 4}\leftarrow$ crop \& rotate $tensor_{h \times w \times 4}$
  \STATE $\alpha_{cnn}$ $\leftarrow$ CNN($tensor_{13 \times 13 \times 4}$)
  \STATE $cp_{new}\leftarrow$ $pixelstep_1 (\alpha_{cnn})$
  \STATE $\alpha_{new} \leftarrow cp_{new}$
  \ENDFOR
 \RETURN $cp_{new}, \alpha_{new}$
 \end{algorithmic}
 \label{algoWtL}
\end{algorithm}
Algorithm \ref{algoWtL} describes our approach via pseudocode. The original image contains useful information, such as edges and orientations and is one part of the input. To follow the object contour, the predictions of an object contour detector are very helpful and therefore the other part of the input. Here, we use the recently published RefineContourNet (RCN)\cite{10.1007/978-3-030-29888-3_20}. The image and the soft contour map are concatenated to an input tensor of size $h\times w\times 4$, where $h$ is the height, $w$ is the width of the image and $4$ channels are obtained through the concatination. Our contour tracer needs only small patches from this tensor to operate. To find and prepare these patches, a center pixel $cp$ and a direction angle $\alpha$ are defined. For initialization, the tracer is either placed on or directly next to the object contour. This starting pixel is selected by choosing a high probability value from the soft contour map and assigning the coordinates to the variable $cp$. The direction angle $\alpha$, describes the direction of the contour at this point. The $cp$ is used to crop the image and the soft contour map at the correct location in which the CNN operates. The first cropping is done quite generously, so that enough outer lying pixels can be considered for the rotation calculation. Then, $\alpha$ is used to rotate this tensor in such a way that the CNN input is always orientated more or less to the direction of the contour itself. This should be benefitting valid predictions, because it provides a similar view for the CNN on upcoming contour lines. A second cropping around the $cp$ brings the tensor to the specified input size of $13 \times 13 \times 4$. This size was chosen because it is approximately the area that allows to make meaningful interpretations without processing too much information. We assume that CNNs are very appropriate for contour tracing, because their convolutional filters are able to learn orientations. Their adaptability and generalization are also well suited for this task. Hence, we have chosen a standard regression CNN as centrepiece. Its architecture is visualized in Table \ref{table_example}. 
\begin{table} 
\renewcommand{\arraystretch}{1.3}
\caption{CNN architecture}
\label{table_example}
\centering
\begin{tabular}{|c|c|c|c|}
\hline
Id. & Layers & Kernel Size & Output Size\\
\hline
1 & Input Layer & - & 13x13x4\\
\hline
2 & Conv1, BN, ReLU & 3x3x4x64 &  13x13x64 \\
\hline
3 & Conv2, BN, ReLU, Pool & 3x3x64x128 & 6x6x128\\
\hline
4 & Conv3, BN, ReLU, Pool & 3x3x128x256 & 3x3x256 \\
\hline
5 & Conv4, BN, ReLU & 3x3x256x512 &  3x3x512 \\
\hline
6 & Conv5, BN, ReLU & 3x3x512x1024 &  1x1x1024 \\
\hline
7 & FC, MSE & 1x1x1024x1 & 1 \\
\hline
\end{tabular}
\end{table}
An input layer feeds the CNN with the cropped and rotated input. Following layers, such as Convolution (Conv), Batch Norm (BN), Rectified Linear Unit (ReLU), Pooling (Pool) and Fully Connected (FC) estimate a final prediction of one single value. All convolutional layers have a stride of 1 and use padding to preserve the input resolution, except for the last convolution layer, which does not use padding. For pooling we used maximum pooling. The Mean Squared Error (MSE) loss is optimized by the standard Stochastic Gradient Descent (SGD) with momentum. The CNN output predicts the following contour course from the $cp$ in clockwise direction and has values between $\ang{-180}$ to $\ang{180}$. With this result a new center pixel $cp_{new}$ and a new input can be found and prepared. This input leads to another new direction estimation and this procedure can be repeated as required for $x$ steps. A possible course of the WtL algorithm is shown in Fig. \ref{onetracer} for a ship image.
\begin{figure*}
\centering
\subfloat[Original \cite{unsplash}]{\includegraphics[width=2.75in]{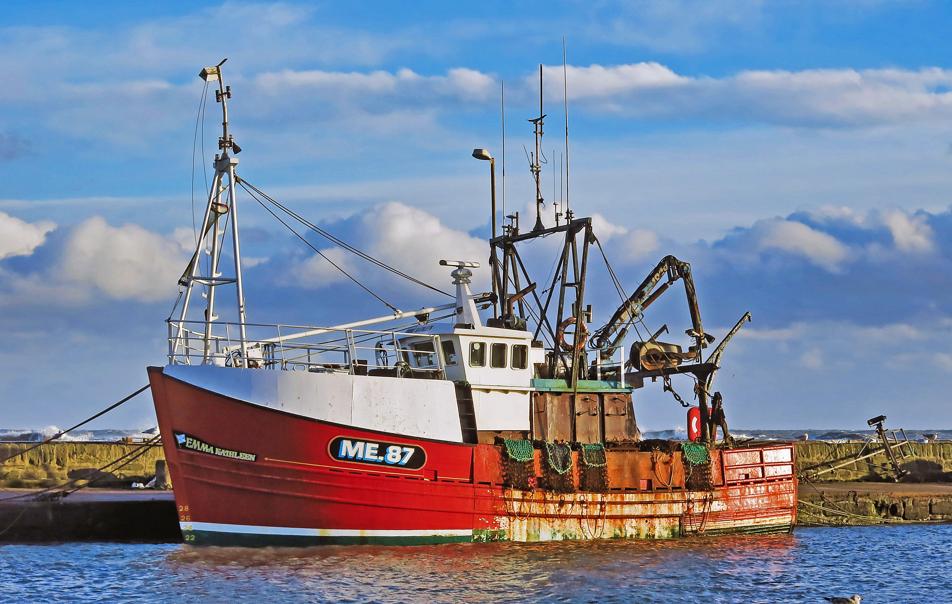}%
\label{fig_first_case}}
\hfil
\subfloat[Ground truth]{\includegraphics[width=2.75in]{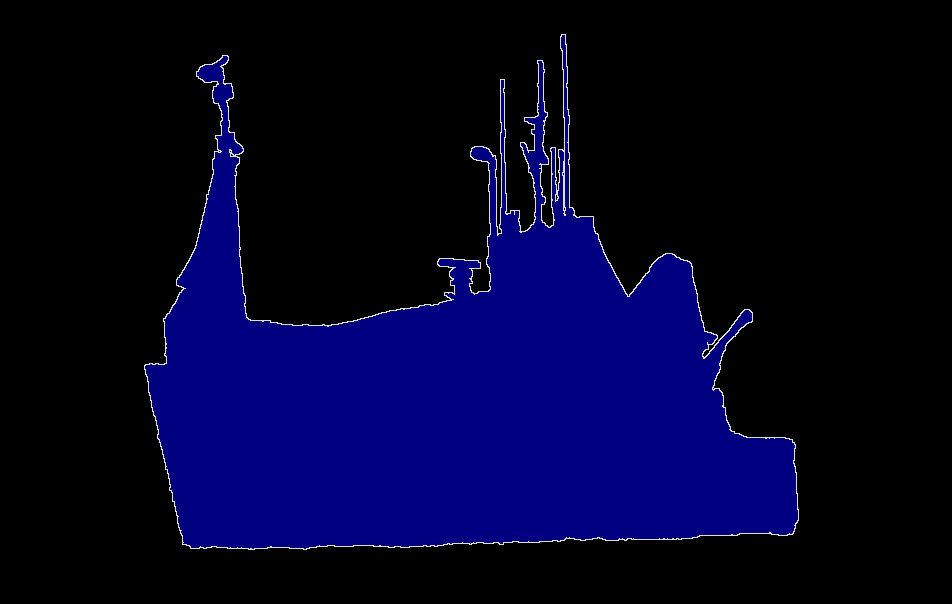}%
\label{fig_second_case}}
\caption{Example of the Detailed Ship Contour (DSC) Dataset.}
\label{DSC}
\end{figure*}

\subsection{Training}
In order to train a regression CNN to follow contours, training data with very detailed contours is required. Since we were not able to find anything in this level of detail, we have created our own dataset. To keep the required number of ground truths manageable and to test the principle first, we focus on one object category, here ship. We create 100 ground truths (gt) with help of the Pixel Annotation Tool \cite{Breheret:2017} and call this internal collection Detailed Ship Contour (DSC) dataset. An example is given in Fig. \ref{DSC}. It shows that great importance is attached to antennas and ship superstructures. Its format is exactly the same as the image segmentation dataset from PASCAL VOC \cite{pascal-voc-2012} and contains an unlabelled region around the ship segmentation mask. Here, however, this region is a line, which has a pixel width of 1 and is identical to the object contour of the ship. In the following, we describe in detail how we generate training labels from the DSC for the CNN.
To have an orientation on the contour gt, we define that the tracer will move around the object in a clockwise direction. A random point on or directly besides the contour gt is chosen. Furthermore, the contour gt is followed by three pixels. This corresponds to the direction of the contour gt at this position, which is noted as the angle $\alpha_0$ for the rotation. The reached pixel gets defined as the center pixel $cp$. This in turn leads to a generous crop of the stacked input, which consists of the image and its matching soft contour maps of the RCN. This is rotated according to the angle $\alpha_0$ and cropped again for the fixed CNN input size of $13 \times 13 \times 4$. Afterwards, the contour gt is followed by three pixels again, which gives us the coming change in direction on the contour itself and this angle $\alpha_{label}$ is saved as the label value for the $cp$ and the already saved corresponding crop. In total we create $1000$ labels per image and use $90\,\% $ for the training and $10\, \%$  for validation.

\section{Object Contour Completion with WtL}
The example in Fig. \ref{onetracer} shows that the tracer can already follow the object contour for a certain distance. But a single tracer has many possibilities to deviate from the correct object contour and so it is quite rare for a tracer to circle the entire object on its own. Therefore, we want to use many tracers at different image locations, which together should be able to draw a complete object contour.

\begin{figure}
\centering
\includegraphics[scale=0.75]{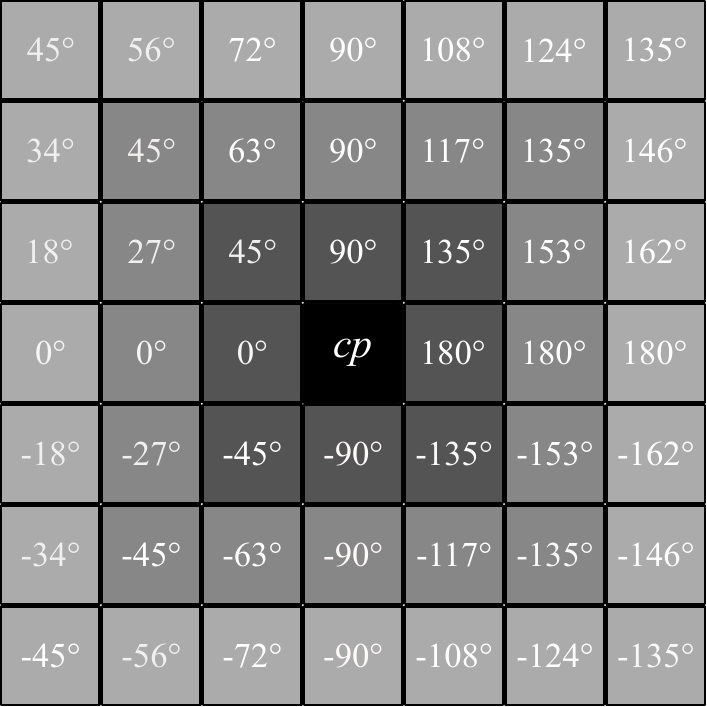}
\caption{Possible directions $\alpha_{new}$ for the three different pixelsteps from the perspective of the center pixel $cp$.}
\label{Pixelsteps}
\end{figure}

\begin{figure}
\centering
\includegraphics[trim=0 15 0 35, clip, width=3.35in]{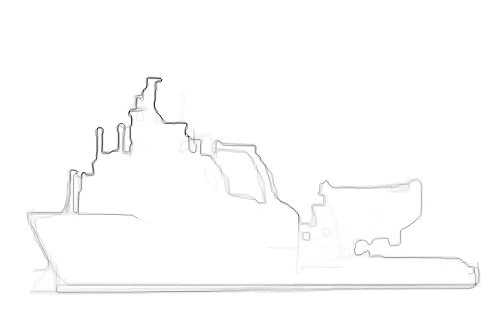}
\caption{The summation of all walked lines gives the WtL contour, visualized here as an example for the same image as in Fig. \ref{onetracer}.}
\label{WtLcontour}
\end{figure}

\begin{algorithm}
 \caption{\textit{Object Contour Completion with WtL}}
 \begin{algorithmic}[1]
 \renewcommand{\algorithmicrequire}{\textbf{Input:}}
 \renewcommand{\algorithmicensure}{\textbf{Output:}}
 \REQUIRE $tensor_{h \times w \times 4}$
 \ENSURE  WtL contour
 \\ \textit{Initialisation}: $tracers(cp,\alpha_0)$ $\leftarrow$  soft contour map
  \STATE $N_{0} \leftarrow$ No. of $tracers$
  \WHILE {$N > 0$}
  \FOR {$tracer_{1}$ to $tracer_{N}$} 
  \STATE $\alpha_{cnn}$ $\leftarrow$ \textit{Walk the Lines} $(tracer_{n}(cp, \alpha))$
  \STATE $cp_{new}\leftarrow$ $pixelstep_{1,2,3} (\alpha_{cnn})$
  \STATE $path_{n}\leftarrow$ save $cp_{new}$
  \IF {$(cp_{new}=bad)$ \textbf{or} $(tracer_n$ is looping$)$}
  \STATE delete $tracer_n$
  \ENDIF
  \ENDFOR
  \ENDWHILE
  \STATE WtL contour $ \leftarrow \sum\nolimits_{n=1}^{N_{0}} path_{n}$
 \RETURN WtL contour
 \end{algorithmic}
 \label{contourenhanceWtL}
\end{algorithm}
The implementation of the contour completion is described via pseudocode in Algorithm \ref{contourenhanceWtL}. The input data is the same as for  Algorithm \ref{algoWtL}. For initialization, the soft contour map is binarized. A relatively high threshold 
ensures that only regions that contain the actual object contour are taken into account. The resulting coarse object contour is thinned out and then fragmented by multiplying with a checkerboard pattern, which creates many small 1 pixel wide lines. The end points of these lines serve as starting center pixels for the WtL tracers. The starting direction angles can also be determined for every starting point. A list of $N_{0}$ tracers is created and a loop runs as long as it contains at least one tracer. For every listed tracer the WtL is performed. To save computational time, the CNN is fed batchwise and returns a vector $1 \times N$ with all new direction angles. A pixelstep is chosen randomly to move the tracers 1, 2 or 3 pixels at once. This offers two advantages:
\begin{itemize}
\item A higher pixelstep can represent more directions, as illustrated in Fig. \ref{Pixelsteps}. With a stepsize of 1, only 8 other pixels can be reached from the $cp$ and therefore only eight directions are available. A step of 2 pixels can display 16 different directions. A pixelstep of 3  has 24 directions. Combined there are 32 different directions. This normally allows each tracer to follow the object contour in a smoother way. 
\item Now, the tracers randomly take different paths, which leads to an increased robustness in the overall result, because previously, many tracers have taken the same wrong turn. 
\end{itemize}
Further robustness is achieved when the pixelstep of 1 is used most often, the one with 2 pixels less often and 3 pixels quite rarely. The percentage of use for these three cases are $87\, \%$, $12\, \%$ and $1\, \%$. The reason for this is that a larger pixel step is less forgiving, because a wrong prediction moves the tracer further away from the contour. The chosen stepsize and the $\alpha_{new}$ move the tracer to the new center pixel $cp_{new}$. If a tracer reaches a bad location it gets deleted from the list. A bad location is when a tracer leaves the image or if predictions of the soft contour map are too unlikely to be an object contour. We also delete it when the tracer is crossing its own old path, then we assume that it is looping or has walked successfully around the object. When no tracer is left, we repeat the whole procedure with the flipped image in order to trace in the anti-clockwise direction. The complete run takes several minutes per image, which results from the pixelwise tracing of the contour. Finally, we sum up all walked lines and return the WtL contour as shown in Fig. \ref{WtLcontour}.

\section{Object Contour Binarization}
\begin{figure}
\centering
\includegraphics[trim=0 15 0 35, clip, width=3.35in]{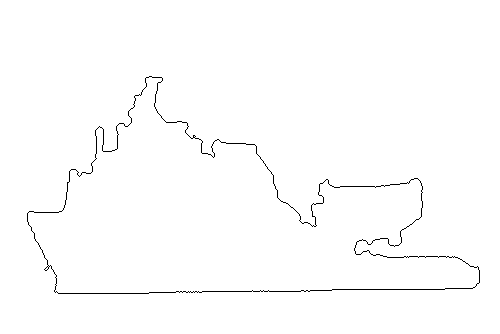}
\caption{Binarization of the WtL contour from Fig. \ref{WtLcontour}.} 
\label{binaryWtL}
\end{figure}

The previously proposed WtL contour is a grayscale image and must now be converted into a binary, 1 pixel wide and detailed object contour. For this, a third procedure is implemented and described in Algorithm \ref{ObjContConmple}. To obtain as much detail as possible, the procedure searches for the highest threshold that closes the object contour. For this purpose, the WtL contour has to be opened at a location which is easy to interrupt. This means when it is certain that no other parts of the object contour will be modified. This is not the case, for instance, in image areas where the contour is not particularly straight. Since only ship images are used, we can use the waterline of the ship. A low threshold of the WtL contour is used to create a binary image, which contains broad contours. There, the longest line is detected, which is very often the waterline. This is done with the Hough transform \cite{DudaH72}. Now a vertical cut is done in the middle, by setting pixels on the cut to zero and copying the former values into a small column called $cutout$. This is performed on the WtL contour and two pixels adjacent to this cut are chosen automatically. The maximum available value in this open WtL contour is defined as the start threshold $th$. With this $th$, a binary image of the open WtL contour is formed. This is checked in a while loop until the $th$ decreasing by $\Delta th$ allows a closed object contour. We assume that the object contour has been closed when the two separated pixels belong to one common segment again. Then the $cutout$ is added back to the WtL contour and a binary image gets created with the determined threshold. The resulting contour is too wide and has branches. A thinning and a cleaning lead to the targeted 1 pixel wide binary object contour, visible in Fig. \ref{binaryWtL}. All algorithms are implemented with MATLAB \cite{MATLAB:2016b} or MatConvNet \cite{vedaldi15matconvnet}.
\begin{algorithm}
 \caption{\textit{Object Contour Binarization}}
 \begin{algorithmic}[1]
 \renewcommand{\algorithmicrequire}{\textbf{Input:}}
 \renewcommand{\algorithmicensure}{\textbf{Output:}}
 \REQUIRE WtL contour
 \ENSURE binary WtL contour
  \STATE longest line $\leftarrow$ WtL contour
  \STATE $pixel_{1}$, $pixel_{2}$, $cutout$ $\leftarrow$ longest line
  \STATE open WtL contour $\leftarrow$ WtL contour$(cutout)$
  \STATE $th \leftarrow$ max$($open WtL contour$)$
  \STATE binary image $\leftarrow$ open WtL contour$(th)$
  \WHILE {binary image$(pixel_{1}$ not connected to $pixel_{2})$}
  \STATE $th \leftarrow th - \Delta th$
  \STATE binary image $\leftarrow$ open WtL contour$(th)$
  \ENDWHILE
  \STATE closed WtL contour$\leftarrow$ $cutout$ $+$ open WtL contour
  \STATE binary image $\leftarrow$ closed WtL contour$(th)$
  \STATE binary contour $\leftarrow$ thinning, cleaning of binary image
 \RETURN binary WtL contour
 \end{algorithmic}
 \label{ObjContConmple}
\end{algorithm}

\begin{table*}
\renewcommand{\arraystretch}{1.3}
\caption{Results on 10 DSC validation images sorted by highest IoU}
\label{table_results}
\centering
\begin{tabular}{|c|c|a|a|c|c|c|a|c|a|c|c||c|a|}
\hline
Method & Metric           & 1     & 2     & 3     & 4     & 5     & 6     & 7     & 8     &  9    & 10 &  $\varnothing $ & $\varnothing_{1, 2, 6, 8} $\\ \hline
\multirow{ 3}{*}{WtL} & P & 94.23 & \textbf{93.66} & 77.04 & 91.90 & 84.11 & 86.88 & 72.67 & 82.56 & 73.10 & 91.47 & 82.66 & 88.54 \\
                      & R & \textbf{99.60} & \textbf{99.05} & \textbf{99.98} & \textbf{98.99} & \textbf{99.88} & \textbf{99.82} & 50.13 & \textbf{99.25} & 72.13 & 11.65 & 50.13 & \textbf{99.52}\\
                      & IoU & \textbf{93.87} & \textbf{92.83} & 77.03 & 91.04 & 84.02 & 86.75 & 42.17 & 82.05 & 57.00 & 11.52 & 70.36 & \textbf{88.17}\\ \hline
\multirow{ 3}{*}{RN} & P & \textbf{98.52} & 87.76 & \textbf{95.97} & \textbf{97.32} & \textbf{96.68} & \textbf{96.52} & \textbf{95.07} & \textbf{93.37} & \textbf{87.64} & \textbf{96.86} & \textbf{94.55} & \textbf{95.00}\\
                      & R& 88.27 & 94.16 & 95.34 & 93.92 & 92.93 & 90.86 & \textbf{88.69} & 89.63 & \textbf{93.68} & \textbf{78.41} & \textbf{90.98} & 90.25\\
                      & IoU & 87.11 & 83.23 & \textbf{91.67} & \textbf{91.55} & \textbf{90.06} & \textbf{87.98} & \textbf{84.79} & \textbf{84.26} & \textbf{82.75} & \textbf{76.46} & \textbf{86.45} & 86.16\\                       
\hline
\end{tabular}
\end{table*}
\section{Evaluation}
\subsection{WtL contours}
\begin{figure*}
\centering
\subfloat[Original \cite{pascal-voc-2012, unsplash, FI}] {\includegraphics[width=1.675in]{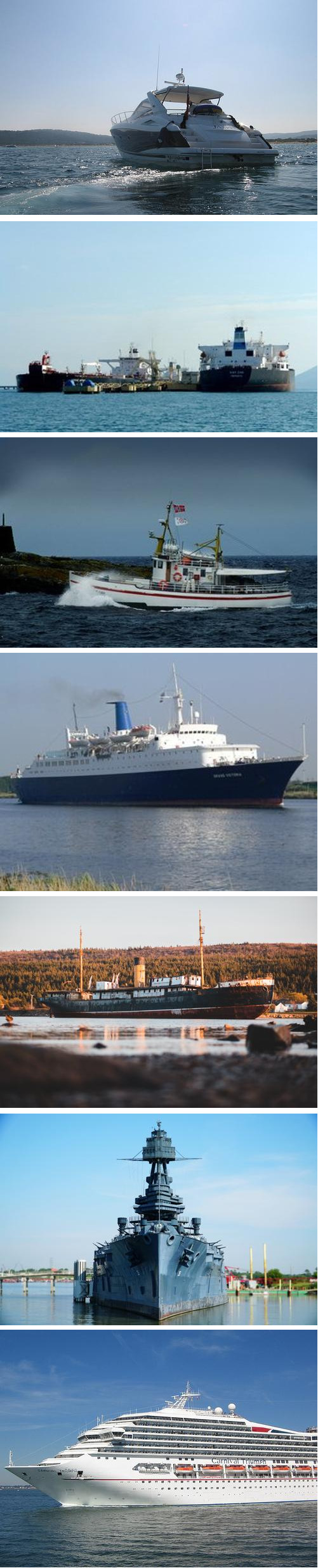}%
\label{rgbs}}
\hfil
\subfloat[RCN]{\includegraphics[width=1.675in]{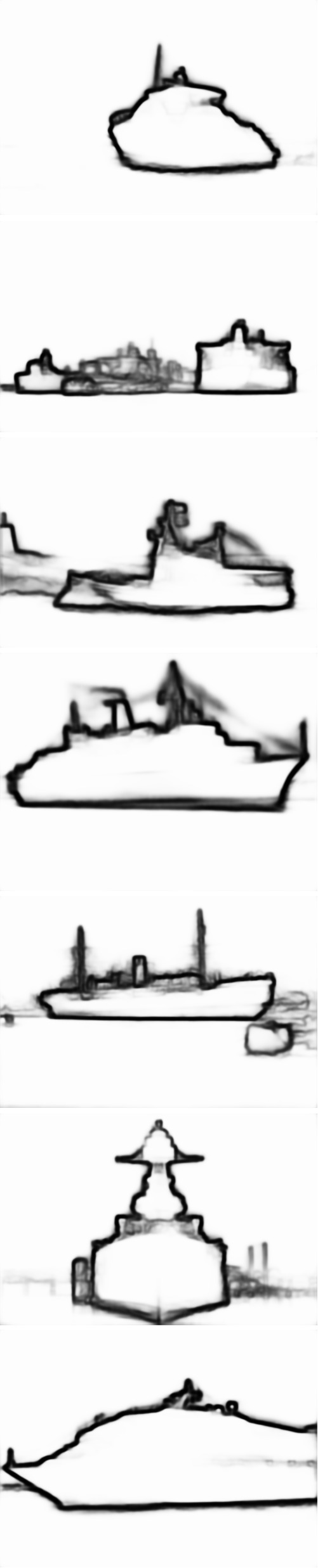}%
\label{objectcontour}}
\hfil
\subfloat[NMS]{\includegraphics[width=1.675in]{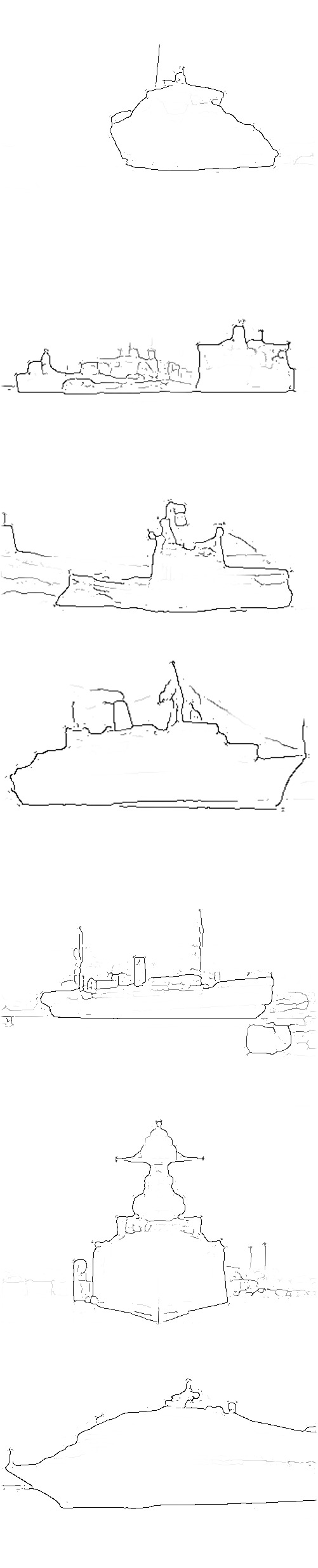}%
\label{NMS}}
\hfil
\subfloat[WtL contour]{\includegraphics[width=1.675in]{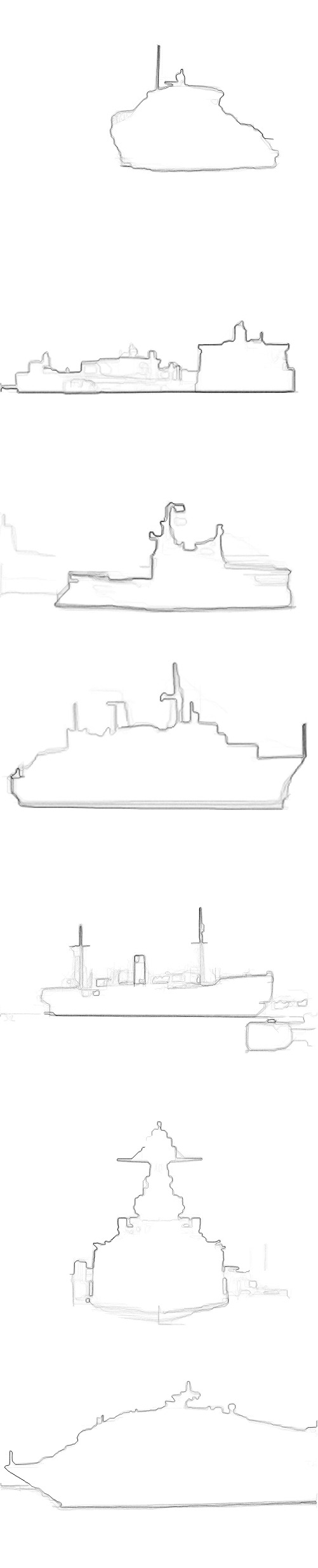}%
\label{wtL}}
\caption{Visualization of the original image (a), the RCN (b), its further processing by NMS (c) and the WtL contour (d) for ship images.}
\label{results}
\end{figure*}

Figure \ref{results} shows the soft object contour map of the RCN and its further processing with the NMS compared to our WtL. Some details are suppressed by the WtL, for instance the second flag at the lifeboat at row 3. Overall, however, more details are worked out. Well visible in the 4th row, the chimney of the passenger ship is closer to the shape of its original. For each individual image, it is visible that our algorithm, in contrast to the NMS, does not break the object contour, while, creates a connected one, instead. This is also true for the yacht in the top image, where the NMS separates the antenna from the ship, while it remains connected when using WtL. The NMS thins the contours to the desired width of 1 pixel, while the WtL sometimes creates wider or parallel contours, clearly visible at the waterlines, which require further processing.
\subsection{Binary WtL contours} 
\begin{figure*}
\centering
\subfloat[Original \cite{pascal-voc-2012, unsplash, FI}] {\includegraphics[width=1.675in]{figs/ICPR_WtL_RGB.jpg}%
\label{rgbs2}}
\hfil
\subfloat[Binary WtL]{\includegraphics[width=1.675in]{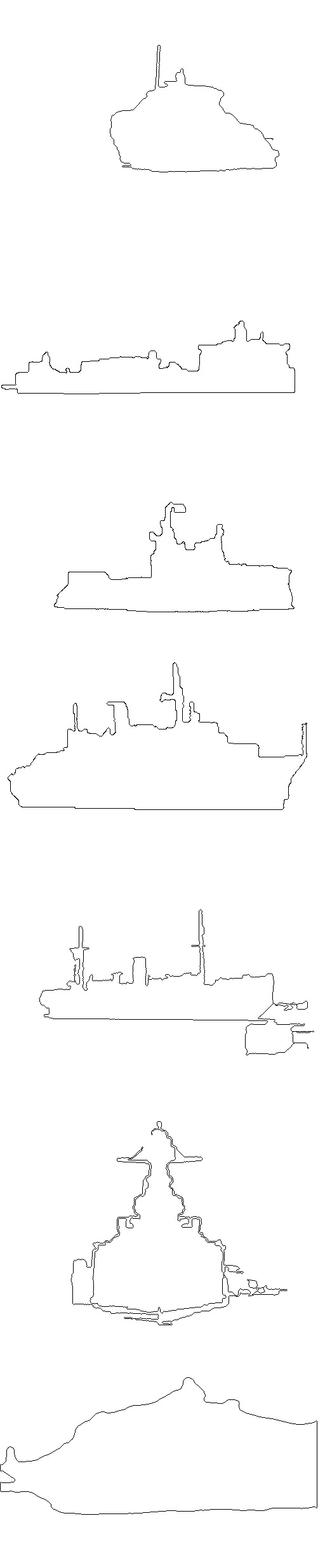}%
\label{objectcontour2}}
\hfil
\subfloat[Segmentation by RN]{\includegraphics[width=1.675in]{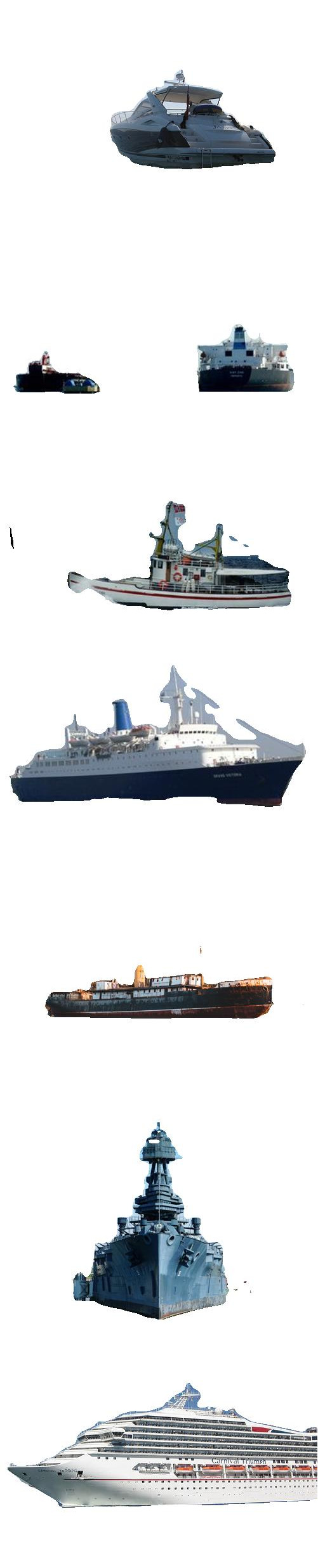}%
\label{NMS2}}
\hfil
\subfloat[Segmentation by binary WtL]{\includegraphics[width=1.675in]{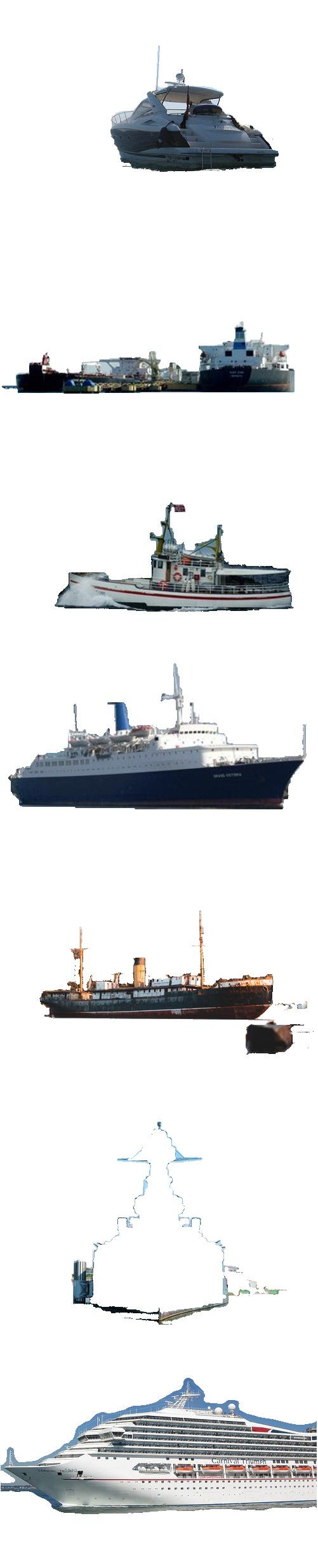}%
\label{wtL2}}
\caption{Visualization of the original image (a), the binary WtL contour (b), the segmentation by RN (c) and the segmentation by the binary WtL contour (d) for ship images.}
\label{results2}
\end{figure*}

For a comparison, we trained RefineNet (RN) \cite{RefineNet} with parameters comparable to those used in RCN. This RefineNet is specialized on the Ship Scene Segmentation and has strong results on PASCAL val2012 with an Intersection over Union (IoU) of $82.5$ for ships, which is comparable with state-of-the-art deep learning segmentation methods. For evaluation, we run the algorithms on the $10$ remaining validation images from the DSC. With the closed contour, we can cut out a segmentation mask and compare it with the ship segmentation mask from the ground truth and calculate the Precision (P), Recall (R) and IoU for each image in Table \ref{table_results}. The advantages and disadvantages are visible here and we have sorted the images by the IoU. On the right are the images where the WtL did not work so well, i.e. where, for example, only a relatively low threshold produced a closed contour resulting in a bad IoU. Another difference between the approaches lies in the different distribution in precision and recall. While the RN has a higher precision and therefore has fewer false positives, the WtL achieves very high recall values, which shows that the WtL has fewer false negatives for some images. This is because the WtL tends to form an outer object contour as it circles the object, which encloses many fine details of the ship, but also false positives. The advantage of our proposed approach is visible on the left side. Here, the contour closing happens quickly by finding a relatively high threshold for that and these columns are marked grey-shaded in Table \ref{table_results}, resulting in an outstanding IoU and a very detailed binary object contour. Because our DSC dataset remains internal, visual results are shown on publicly available ship images in Fig. \ref{results2}. The images in the last three rows show three typical examples where the binary contour does not give satisfactory results:
\begin{itemize}
\item "Wrong Lines" include unwanted pixels to the segmentation, visible at the stranded ship, where the stone in the foreground is wrongly segmented.
\item "Doubled Lines" appear, if the contour closing does not pass over the desired object contour. This results in a poor segmentation, visible at the warship.
\item "Open Lines" leads to a very low threshold, because a closed contour is found very late. The result is a high loss of detail, as can be seen on the passenger ship, the image at the bottom.
\end{itemize}
Extracting the binary object contour is not guaranteed, because it relies on a chain of previous procedures, such as an accurate RCN prediction or a successful object contour completion. In addition, assumptions are made for the binarization step which do not necessarily apply. This includes the assumption that the longest line is really the waterline of the ship. Therefore, we can only give an approximate range of $20\, \%$ - $45\, \%$ for the creation of an acceptable binary WtL contour, based on our experience and always depending on the individual image. The advantage of our method is only visible for those images on which the whole procedure works as desired. Then, excellent segmentations with a very high level of detail become possible, as visible in the upper four images in Fig. \ref{wtL2}. Compared to the RN in Fig. \ref{NMS2}, more details of the ship superstructures and thin antennas are visible in the ship segmentation.

\section{Conclusion}
The WtL shows that contour tracing is a new and unexplored application for CNNs.
The object contour completion by WtL draws excellent object contour maps on which many specific details of the original object can be seen. Compared to the results from the NMS, the contour is a bit more detailed and wider. The main advantage, however, is that WtL contours are connected and can be converted into a closed binary contour without losing many details. The binarization is completely automatic, but so far it only works for a limited number of images. For these we produce excellent segmentations with very high IoUs and reveal details that are easily omitted, such as antennas and ship superstructures.

%
%
%

\bibliographystyle{IEEEtran}
\bibliography{IEEEabrv,egbib}

\end{document}